\documentclass[11pt,a4paper]{article}

\usepackage{amsmath,amssymb}
\usepackage{graphicx}
\usepackage{booktabs}
\usepackage{bm}

\title{Beyond Backpropagation: Monte Carlo Method Can Train Deep Neural Networks}
\author{Hong Zhao$^{1,2}$}
\date{}

\begin{document}

\maketitle

\noindent
$^1$ Department of Physics, Xiamen University, Xiamen 361005, China\\
$^2$ Lanzhou Center for Theoretical Physics, Lanzhou University, Lanzhou 730000, China\\
Email: zhaoh@xmu.edu.cn

\begin{abstract}
Backpropagation (BP) dominates deep learning training, but its reliance on gradients brings inherent troubles---vanishing and exploding gradients. The pursuit of gradient-free methods has long been a goal in the field of artificial intelligence. This paper shows that indeed the simplest Monte Carlo algorithm implemented on a single GPU---randomly mutate a parameter, keep it if the loss decreases, otherwise retry---can practically train deep networks. This gradient-free method does not even need common techniques such as batch normalization or residual connections to directly train sufficiently deep networks. More remarkably, its flexibility extends to several nontrivial scenarios: it enables pure pruning training, supports discrete weights, accommodates unconventional transfer functions such as Gaussian, and reveals the substantial redundancy of deep networks. We have demonstrated its feasibility on deep networks with more than 20 layers, single-hidden-layer wide networks with up to 16,384 hidden neurons, and even a simple Transformer architecture trained on both image classification (MNIST) and character-level language modeling (Tiny Shakespeare). This simple gradient-free method may offer a complementary perspective for understanding the self-organization and learning mechanisms of neural networks, and also provides an alternative route for building physically inspired deep learning systems.
\end{abstract}


\section{Introduction}

Although backpropagation (BP) is currently the main method for training learning machines~\cite{Rumelhart1986,LeCun2015}, it is still necessary to explore alternative algorithms. Representative works in this field include random forests~\cite{Ho1995,Breiman2001,Scornet2026}, kernel methods~\cite{Scholkopf2002,Du2024}, tensor network architectures and their corresponding algorithms~\cite{Novikov2015,Zhangp2018,Zhangp2022}. In recent years, increasing attention has been paid to finding new paradigms that can replace or even surpass BP. For example, spiking neural networks have explored brain-like learning~\cite{Neftci2019,Bellec2020,Stanojevic2024}, and the local-learning-based Forward-Forward algorithm has attempted to directly replace BP~\cite{Hinton2022}. However, these methods have so far been unable to challenge the dominant position of BP in mainstream deep learning.

The Monte Carlo method is a fundamental computational tool in physics, successfully applied to a series of key problems in statistical physics, quantum field theory, particle transport, and other fields~\cite{Metropolis1953,Landau2021,Creutz1980,Agostinelli2003,Tramm2024}. By introducing stochastic sampling, it breaks through the limitations of traditional deterministic prediction models and has been widely used in various branches of neural networks~\cite{Neal1996,Andrieu2003,Chandra2024}. Among them, Monte Carlo Dropout is widely used in various deep networks~\cite{Gal2016,He2026}, and diffusion models are built upon Monte Carlo sampling~\cite{Song2021}. Moreover, the application of Monte Carlo tree search is the key to the success of top artificial intelligence systems such as AlphaGo~\cite{Silver2016}.

In the early days of artificial neural network research, the Boltzmann machine had already introduced the concepts of random sampling, thermal equilibrium distribution, and simulated annealing from statistical physics into the basic architecture and learning rules of neural networks~\cite{Ackley1985,Hinton1986}; however, because such methods require repeated operation of the network to estimate equilibrium statistics, the learning process is very slow, and they therefore long remained a non-mainstream training paradigm~\cite{Ackley1985,Hinton2007}. More than twenty years ago, this method was used to design asymmetric Hopfield networks~\cite{Zhao2004}, which have advantages over symmetric networks, such as controlling the proportion of spurious attractors and increasing storage capacity~\cite{Zhao2004,Jin2005}, and storing sequential information in the form of limit cycle attractors~\cite{Wu2008}. Subsequently, the method was extended to the training of general neural networks and shallow fully connected networks~\cite{Zhao2006,Xin2009,Zhao2017,Zhao2021}: by designing an appropriate algorithm that calculates only the changed parts caused by mutations rather than evolving the entire network, training of shallow fully connected networks became feasible on a CPU. Its flexibility allows direct constraint of the magnitude range of each weight, thereby effectively controlling generalization ability~\cite{Zhao2017}. This algorithm is called the Monte Carlo mutation--optimization selection method (MCA). In addition, related work has applied MCA to discrete-weight feedback networks, diluted neural network dynamics, and network spectral structure analysis, further demonstrating the strong flexibility of this method in network structure design and dynamical regulation~\cite{Wang2013,Zhou2009,Wang2016}. However, limited by the computational efficiency of CPUs, the above methods are mainly confined to shallow or small-scale network training, and have not yet been able to use the Monte Carlo method as a basic training tool for designing deep networks.

This paper aims to reveal that, on a GPU framework, the most basic Monte Carlo algorithm can serve as a practical tool for designing deep networks and can serve as an effective complement and extension to the BP algorithm. We first present the specific implementation and numerical experiments, including demonstrations that single-parameter MCA operations, pure pruning (weight-zeroing), and multi-parameter MCA operations can all effectively design deep networks. We also introduce shallow wide networks designed with GPU parallelization and demonstrate the feasibility of training Transformer architecture networks with the MCA method on both image classification (MNIST) and character-level language modeling (Tiny Shakespeare). Then we provide a theoretical analysis comparing some properties of MCA and BP.

\section{Numerical Results}

The basic MCA algorithm is as follows: randomly select one (single-parameter mutation) or a group (multi-parameter mutation) of network parameters (which may include weights, biases, or transfer function coefficients), and apply a random operation:
\begin{equation}
\begin{split}
w &\rightarrow w + \delta \cdot \textit{rand}, \\
&\qquad \text{accept the mutation if } \text{loss}(\text{new}) \leq \text{loss}(\text{old}) \text{ and } |w| < \Delta,
\end{split}
\end{equation}
where \textit{rand} is a uniform random number in the interval $[-1,1]$, $\delta$ controls the perturbation amplitude and is called the learning rate, and $\Delta$ is a preset constant to constrain the range of parameter values and control the generalization characteristics of the network. When $\delta \cdot \text{rand} = -w$, the operation is pruning (weight zeroing); if $w$ is taken as a discrete value so that the mutation only jumps between adjacent discrete states, a discrete-weight network is obtained.

The loss function is the average cross-entropy. All calculations were performed on a single L40S or A100 GPU using programs written in CUDA Fortran. Unless otherwise stated, all numerical computations in this paper are performed in single precision. The numerical experiments in this paper use the standard MNIST dataset (60,000 training samples, 10,000 test samples). We study two typical neuron transfer functions. One is the ReLU function, currently widely used in the training of various AI architectures, defined as $f(x) = \max(0,x)$. The other is the Gaussian function, of the form $f(x) = \exp(-\gamma x^{2})$ ($\gamma>0$). Because this function is unfriendly to BP, it is rarely used directly in deep networks.

\begin{figure}[htbp]
\centering
\includegraphics[width=8.0cm]{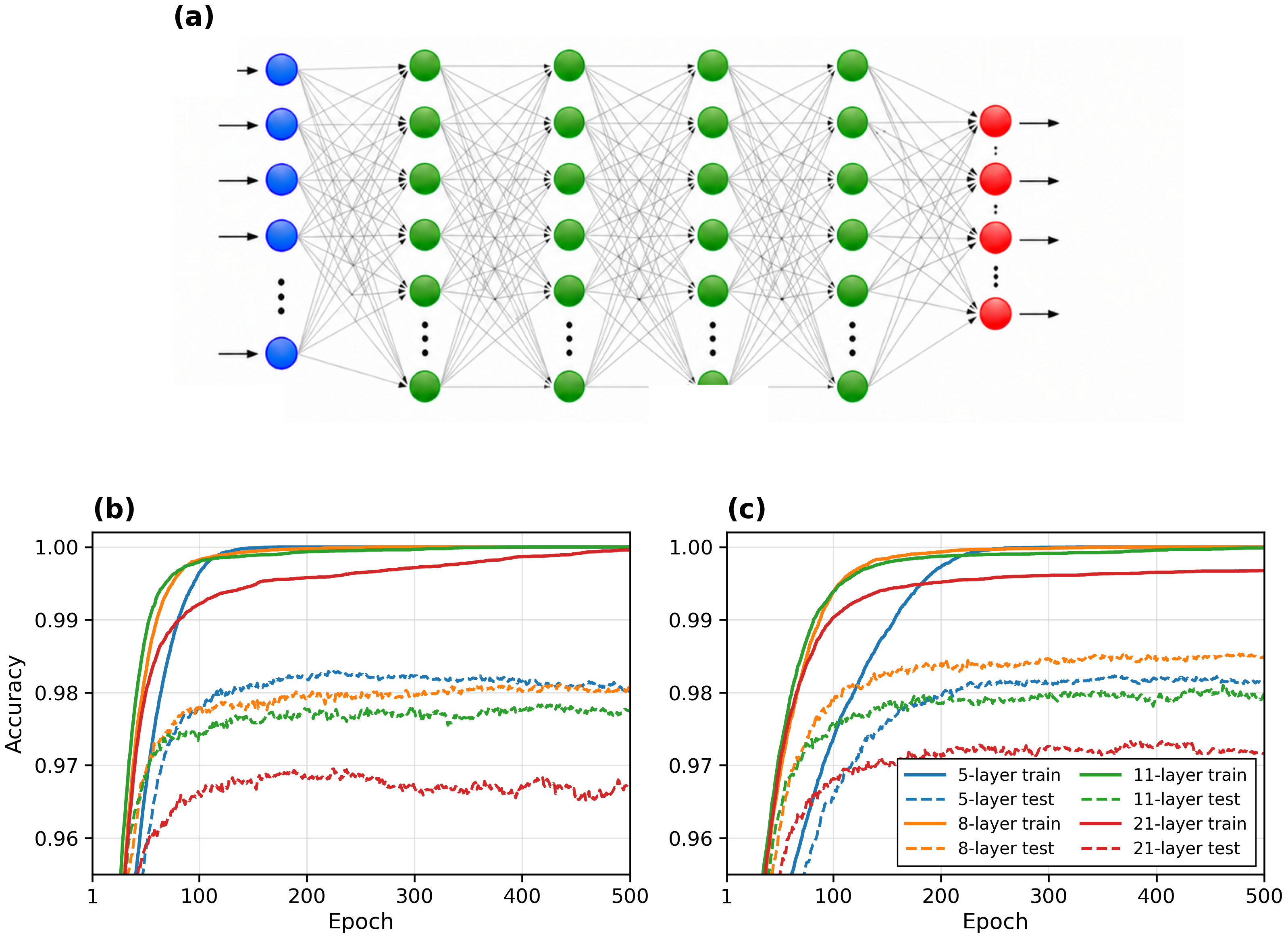}
\caption{Fully connected deep neural networks trained by MCA. (a) Schematic of a representative fully connected feedforward network. (b) Training and test accuracy versus epoch for networks with 5, 8, 11, and 21 layers using ReLU activation. (c) Same as (b) but with Gaussian activation. Solid lines: training accuracy; dashed lines: test accuracy.}
\label{fig1}
\end{figure}

\subsection{Single-parameter MCA Training of Deep Networks}

Figure~\ref{fig1}(a) illustrates a typical fully connected multilayer deep network structure. To avoid overfitting, network weights are initialized as small random numbers: $w \sim \sqrt{0.1/n} \cdot \text{rand}$, where $n$ is the hidden layer width. The parameter constraint amplitude is $\Delta = 0.1$, and the learning rate is $\delta = 0.01$. To accelerate deep network training, we adopt a current-layer caching strategy: randomly select a layer, perform forward propagation over all samples, cache the local field data of the preceding hidden layer, then perform multiple Monte Carlo mutations within that layer, move to the next randomly selected layer, and repeat. This avoids recomputing the local fields of layers before the mutated parameter for each mutation, because these data actually do not change.

Figure~\ref{fig1}(b) shows the training accuracy and test accuracy as functions of training time for networks with depths of 5, 8, 11, and 21 layers, each with width 256, under the ReLU activation function. In this paper, 10,000 Monte Carlo operations are defined as one training epoch. The networks in this example use layer normalization (LN), and the batch size is fixed at 60,000 (i.e., full-dataset training). Unless otherwise stated, the full-batch setting is adopted for all numerical experiments in this paper. As can be seen from the figure, the training accuracy quickly reaches 100\%; the test accuracies over the training interval reach 98.30\%, 98.11\%, 97.83\%, and 96.95\%, respectively. For reference, using the standard BP algorithm (Adam optimizer) to train a 5-layer network of the same structure with full-batch training, the test accuracy after 100 epochs (which already gives sufficient training) is about 98.19\%. Thus, the MCA-designed 5-layer network performs on par with the BP algorithm under the full-batch training setting. It should be noted that the focus of this paper is to demonstrate the feasibility of MCA for designing deep networks, rather than performing a fine comparison of extreme accuracy values with BP, which would require tedious parameter optimization for different batch sizes. The accuracy obtained by MCA is related to parameters such as the initial distribution, $\Delta$, and $\delta$. We use a set of parameters ($\Delta=0.1$, $\delta=0.01$) that perform well for a 5-layer 256-width network, and apply them directly to networks of other depths. In practice, there exist optimal parameter combinations for different depths and widths.

From the training progress, the accuracy becomes nearly saturated after 250 epochs. The training time for 250 epochs of the 5-layer network is about 2.5 hours. The training time for the same number of epochs is roughly proportional to the network depth. For comparison, under the same network scale and full-batch setting, a BP network achieves sufficient convergence within 100 epochs, with a training time of about 23 minutes. Thus, although MCA training is slower than BP, it remains within the same order of magnitude.

In particular, by avoiding gradient problems, MCA may circumvent some of the difficulties encountered by BP when applied to other transfer functions. Figure~\ref{fig1}(c) shows the results for the Gaussian function with the same network scales. The initialization and learning rate are the same as for the ReLU case, and the coefficient of the transfer function for each layer is fixed at $\gamma=0.4$. We see that the test accuracies are significantly improved compared to ReLU, reaching 98.23\%, 98.53\%, 98.10\%, and 97.34\%, respectively. For comparison, training a 5-layer Gaussian transfer function network with the BP algorithm (Adam optimizer) achieves only about 97.6\% accuracy. Moreover, the Gaussian results are also significantly better than the MCA algorithm using the ReLU function.

The numbers of parameters for the networks studied here are $3.35\times10^5$, $5.32\times10^5$, $7.30\times10^5$, and $13.9\times10^5$, respectively. Over 500 epochs, a total of $5\times10^6$ Monte Carlo trials are performed. For the 5-layer network, the average number of successful mutations per parameter is less than 3; for the 21-layer network, this number is less than 0.5. This means that the mutation operation does not need to be frequently applied to every parameter. This is in sharp contrast to the BP algorithm, where each parameter is updated 100 times over 100 epochs. This indicates that the MCA algorithm mainly relies on a self-organization mechanism to find mutations that are beneficial to the entire sample set (in a statistical sense) based on the existing structure, and thus does not require all parameters to be updated multiple times, or even updated at all. This is an important difference between the MCA algorithm and the BP algorithm.

\begin{figure}[htbp]
\centering
\includegraphics[width=8cm]{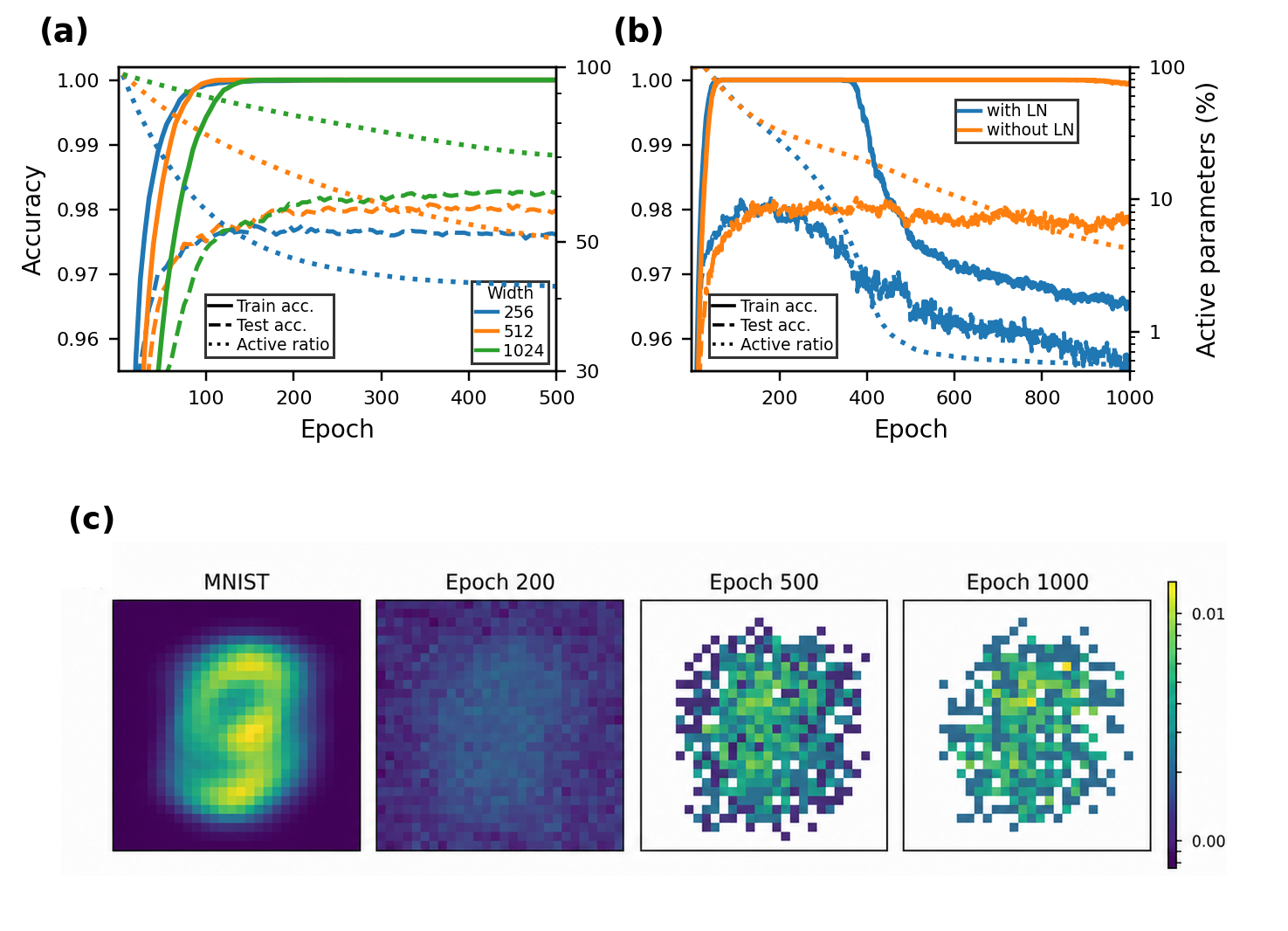}
\caption{Performance evolution of five-layer deep networks during pruning-based training. (a) Training and test accuracy and active-parameter ratio under pure pruning for networks with widths 256, 512, and 1024. (b) Training and test accuracy and active-parameter ratio under mixed pruning and random mutations for width-256 networks, with and without LN. (c) Probability-density maps of MNIST input patterns and the corresponding remaining weight distributions in the input layer at different training epochs, corresponding to the with-LN case in (b).}
\label{fig2}
\end{figure}

\subsection{Pure Pruning Training of Deep Networks}

MCA has high flexibility in training neural networks. For example, it can train using pure pruning operations ($w=0$) or pure weight decay operations ($w\rightarrow w\cdot\delta$, where $|\delta|<1$). Figure~\ref{fig2}(a) shows the results of training five-layer networks with widths 256, 512, and 1024 using pure pruning MCA under the ReLU function. Here, the initial parameter range is given by the standard He initialization $\sqrt{8/n}$ (this initialization is adopted for all neural networks in the subsequent parts of this paper). During training, MCA attempts to set weights to zero, and the pruning is accepted if the loss function does not increase. It should be noted that if the initial weights are set too small, pruning may cause negligible change in the loss function, thereby making it difficult for the network to evolve effectively.

Notably, the training accuracy also reaches 100\%; the test accuracy is slightly lower than that of conventional MCA but does not drop significantly. The accuracy improves with increasing network width. This indicates that deep networks can be trained by a pure redundancy removal operation.

Figure~\ref{fig2}(a) also shows the evolution of the ratio of the number of remaining weights to the initial total number of weights over training time. It can be seen that after 500 epochs, the remaining weight ratios for widths 256, 512, and 1024 are approximately 1/3, 1/2, and 2/3, respectively, indicating that the network achieves effective sparsification.

Under full-batch training, the pruning operation converges to a specific weight distribution after the loss function decreases to a certain level, making further training difficult (because acceptable mutations become very rare). To investigate the maximum pruning rate (i.e., the ultimate sparsity limit) that the network can withstand, we start from the 30th epoch and mix pruning mutations with random weight mutations, prioritizing pruning mutations. At the same time, we use a batch size of 30,000 samples to introduce greater randomness, thereby providing more opportunities for pruning mutations to be accepted. Figure~\ref{fig2}(b) shows the evolution of training accuracy, test accuracy, and pruning rate for a five-layer network with width 256, under two configurations: without layer normalization (LN) and with LN, as training progresses.

It can be seen that without LN, the remaining weight ratio at the end of training is only about 4.0\%, and the test accuracy drops from its highest value of 98.18\% to 97.83\%. With LN, the training accuracy reaches a maximum of 98.17\% at about 200 epochs, then begins to decline around 250 epochs, but the accuracy remains above 95.5\% at the end of training, while the remaining weights account for only about 0.33\% of the total weights of the fully connected network. Specifically, the number of remaining weights drops from 334,336 in the fully connected network to about 1,097, meaning that on average each neuron is connected to less than one weight, demonstrating the extreme sparsity of the pruned network. Using LN causes larger perturbations, especially when the number of weights drops sharply, thereby keeping the pruning operation more active. Nevertheless, even with only about 1,097 weights remaining, the network function does not collapse.

Given that the total number of samples is 60,000, it is a surprising result that the network can withstand such a large degree of pruning. The reduction in weights implies lower operational costs and also indicates that deep networks have very large redundancy. To further reveal how the distribution of remaining weights reflects the way information is extracted during training, Figure~\ref{fig2}(c) shows heat maps of the probability density function of all digit images (28$\times$28 bitmap) in the MNIST dataset, and heat maps of the distribution of remaining weights in the input layer at epochs 200, 500, and 1000. Completely disconnected sites are marked in white. It can be seen that as pruning proceeds, the common background part is gradually removed, while the spatial distribution intensity of the remaining part is basically consistent with the digit region. This indicates that the pruning operation can reveal the way the network encodes information in advance.

\begin{figure}[htbp]
\centering
\includegraphics[width=8cm]{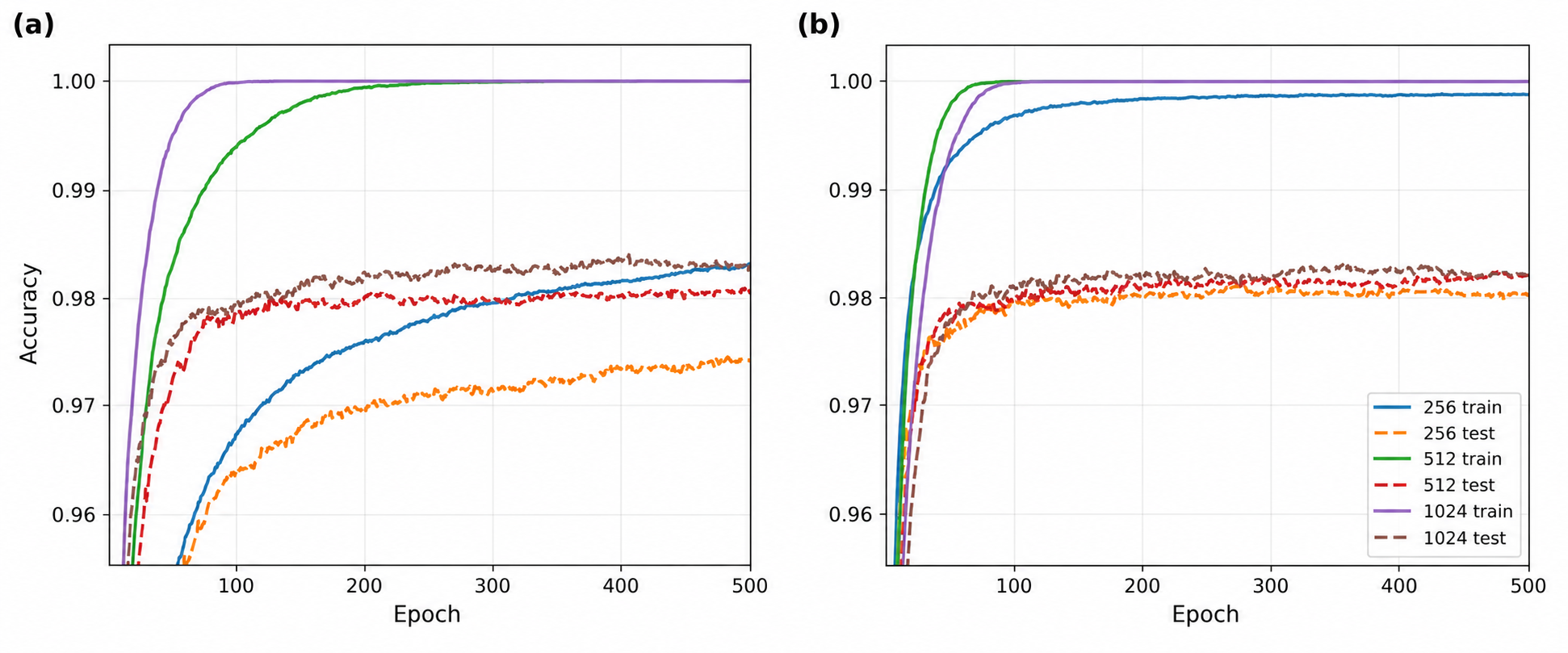}
\caption{Training of deep networks via parameter-channel mutation. (a) and (b) show the evolution curves of training accuracy and test accuracy for three networks of different widths obtained using channel mutation and channel pruning algorithms, respectively, with the ReLU function.}
\label{fig3}
\end{figure}

\subsection{Multi-parameter Mutation Training of Deep Networks}

The basic version of MCA mutates only one parameter at a time, which may limit training efficiency. Exploring simultaneous mutation of multiple parameters may provide a feasible way to increase training speed. Considering that effective parameters should be able to connect input to output, we propose two schemes for training deep networks by mutating a single weight channel from the input layer to the output layer. To simplify the discussion, biases are fixed at zero and layer normalization is not used. A weight channel from input to output is randomly generated, then the first scheme randomly reassigns the weights on that channel with amplitudes normalized by He initial, and accepts the mutation if the loss function does not increase. The second adopts a Dropout-like mutation strategy: all parameters on that channel are set to zero, and the mutation is accepted if the loss function does not increase.

Figures~\ref{fig3}(a) and \ref{fig3}(b) show the results of training five-layer networks with widths 256, 512, and 1024 using these two methods. Compared with Figure~\ref{fig1} (single-parameter MCA), the test accuracy of the multi-parameter mutation networks is lower for the same width. This is due to the fact that when the width is small, the acceptance rate of multi-parameter mutations drops rapidly with training process, making it impossible to effectively reduce the loss function. As network width increases, the acceptance rate increases, and the network accuracy improves accordingly, as shown in the figure.

\subsection{MCA Training of Shallow Wide Networks}

\begin{figure}[htbp]
\centering
\includegraphics[width=7.0cm]{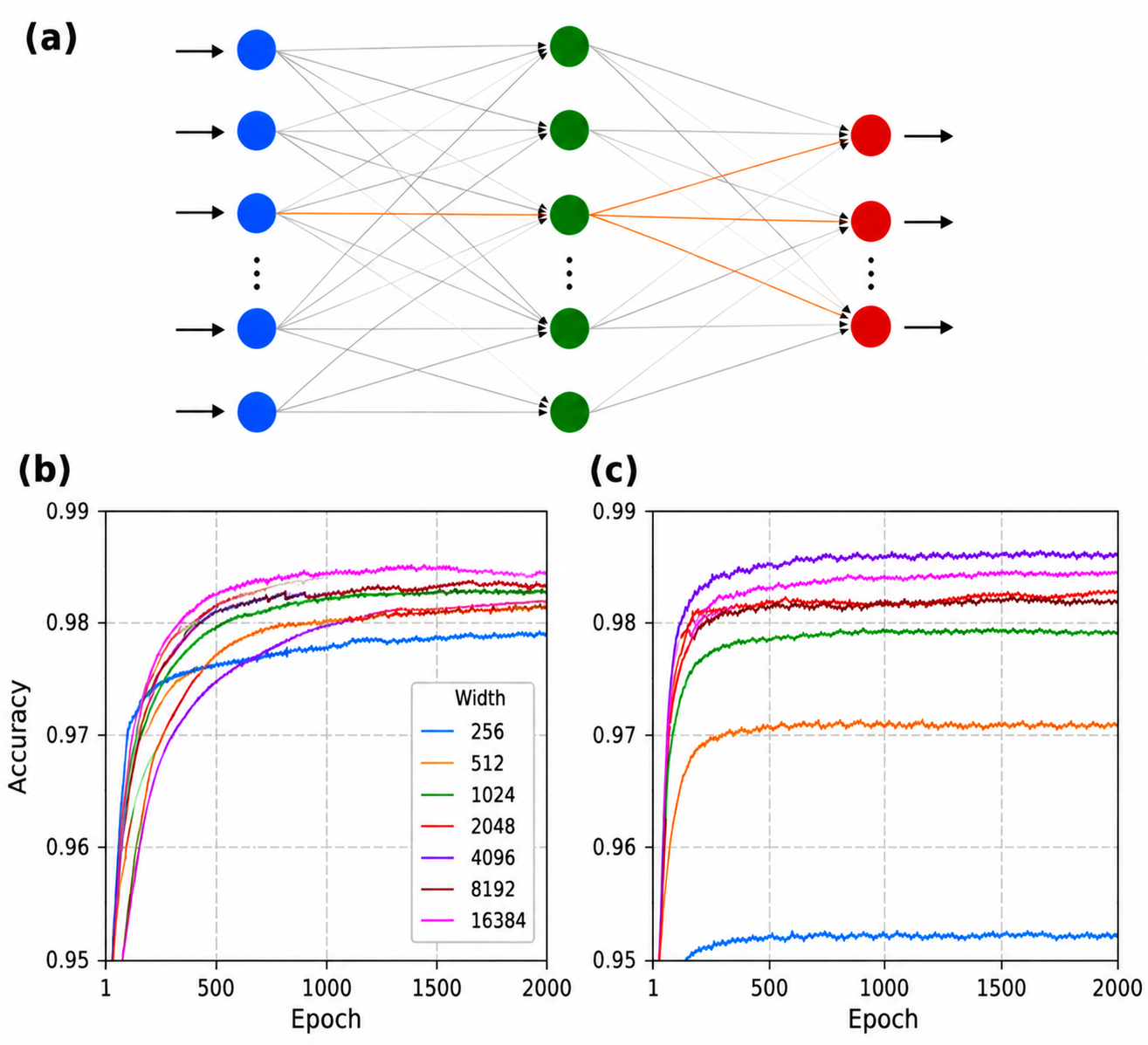}
\caption{Training of single-hidden-layer networks with MCA. (a) Schematic of a single-hidden-layer network, highlighting that mutating a single weight (from input to hidden) only requires recomputing the local fields of the affected neurons (highlighted). (b) Test accuracy versus network width using ReLU activation. (c) Same as (b) but with Gaussian activation.}
\label{fig4}
\end{figure}

Single-hidden-layer networks can be trained on a CPU using MCA, as described in the literature~\cite{Zhao2017,Zhao2021}. The principle is that, using cached data, changing one parameter only requires re-evolving a small part of the network, as illustrated in Figure~\ref{fig4}(a). For example, changing an input-layer weight affects only the input local field of the hidden-layer neuron it connects to, and the local fields of the ten output-layer neurons connected to that hidden neuron, as illustrated in Figure~\ref{fig4}(a).

Based on this characteristic, the GPU architecture can efficiently parallelize these local computations: cache the local field data after the last successful mutation operation, then randomly mutate the next parameter, and decide whether to accept the mutation by computing and comparing the loss function. Figures~\ref{fig4}(b) and \ref{fig4}(c) show the test accuracy as a function of width for ReLU and Gaussian functions, respectively, when the hidden-layer width is increased from 256 to 16384. Learning rate is set to $\delta=0.01$, Gaussian function coefficient fixed at $\gamma=0.04$.

Under the given initialization and control parameters, the ReLU function achieves its highest test accuracy of 98.50\% at width 2048, and the Gaussian function achieves 98.77\% at width 4096. This indicates that MCA is very effective for designing single-hidden-layer wide networks. Under the Gaussian transfer function, the accuracy of wide networks significantly exceeds that of ReLU, but for narrower networks, the Gaussian function exhibits significantly lower accuracy, whereas ReLU is much less sensitive to width. One reason for this is that we set a relatively small value for the Gaussian function, which makes training of narrow networks very slow due to excessively low input--output sensitivity.

Thanks to full utilization of the GPU's parallel computing architecture, the training time is almost independent of width. Each network is trained for 2000 epochs, taking about 38 minutes on an A100 GPU in double-precision mode; the training time for the width-16384 network is only less than 5\% longer than that for the width-256 network. For wide networks, achieving a test accuracy above 98\% takes only about 2 minutes.

The parallel advantage of the GPU can also be used to optimize the training speed of two-hidden-layer networks. For mutations occurring on parameters after the first hidden layer (i.e., from the first to second hidden layer, or deeper), the GPU scheme is the same as for the single-hidden-layer case. If a mutation is chosen on a weight connecting the input layer to the first hidden layer, then for each sample, it will only cause $O(N)$ multiplications in the second hidden layer to be recomputed (due to the fan-out of the affected first-hidden-layer neuron). The output layer still requires a full recomputation, but its cost is low because it connects only 10 neurons (corresponding to the 10 classes of MNIST).

\subsection{MCA Training of Transformer Networks}

\begin{figure}[htbp]
\centering
\includegraphics[width=7.0cm]{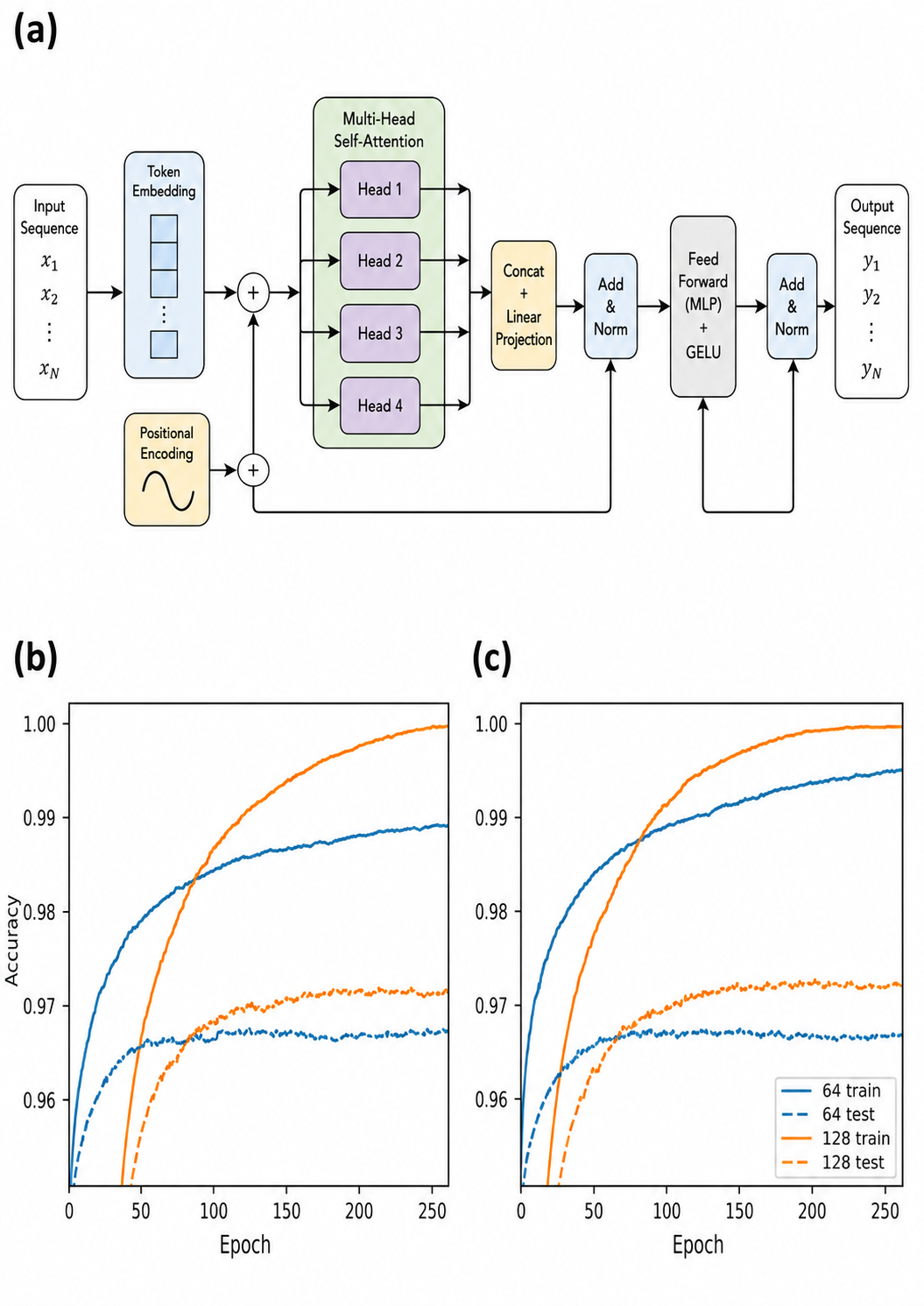}
\caption{Training of Transformer with MCA. (a) Schematic of the four-head Transformer block. (b) Training and test accuracy versus epoch for FFN dimensions 64 and 128 using GELU activation. (c) Same as (b) but with Gaussian activation.}
\label{fig5}
\end{figure}

Figure~\ref{fig5}(a) shows a simple four-head Transformer architecture, where each module is initialized in the conventional way. Unlike fully connected networks, the Transformer contains various types of parameters. Compared with a fully connected network of the same depth, the Transformer architecture has a smaller number of parameters, but its computational cost is much larger. To simplify processing, we select parameters randomly with equal probability for MCA mutations, and the mutation amplitude is limited to the value range corresponding to the initialization of each type of parameter. In this section, we first evaluate the MCA-trained Transformer on MNIST image classification, and then extend the test to a character-level language modeling task (Tiny Shakespeare) as a proof-of-principle demonstration of its generality beyond classification.

Figures~\ref{fig5}(b) and (c) show the training and test accuracy as functions of training time for a three-layer Transformer on the MNIST dataset, with Feed-Forward Network (FFN) hidden dimensions 64 and 128 (Multilayer Perceptron, MLP, dimensions 128 and 512). Panel (b) uses GELU as the transfer function, while panel (c) uses the Gaussian function ($\gamma=1.0$). For both activations, test accuracy exceeds 97\% for the larger configuration, improving with network width. Interestingly, the Gaussian function again shows an advantage, with test accuracy higher than that of GELU. This further suggests that after overcoming the gradient problem, the Gaussian function may become a superior transfer function, at least under certain conditions. Transformer training is, however, considerably slower than fully connected networks: 150 epochs for the FFN dimension 128 configuration require about 48 hours on a single L40S GPU.

The MNIST results above confirm that MCA can train a Transformer on a standard classification task. To further assess its generality on sequence modeling, we apply the same gradient-free procedure to character-level language modeling using the Tiny Shakespeare dataset.

The Tiny Shakespeare dataset, a standard benchmark widely used for testing character-level Transformer language models, was used as a proof-of-principle task for the proposed MCA scheme. The 3-layer model uses a vocabulary size of $65$, a context length of $64$, and three causal Transformer blocks with embedding dimension $d_{\rm model}=64$, $4$ attention heads, and a feed-forward hidden dimension of $256$. It is worth noting that the task here is character-level prediction, rather than token-level prediction---the vocabulary size of $65$ corresponds to the character set of the Shakespeare text, not a tokenized vocabulary. This makes the task substantially smaller in scale than token-level modeling in modern large language models, yet it suffices as a proof-of-principle demonstration of the gradient-free training procedure.

Starting from random initialization, the model was trained for 48 hours with an adaptive batch-size schedule, reaching a final training batch size of $8192$. At the final checkpoint, the model achieved a training loss of $1.650$ and a training accuracy of $0.51$, while the held-out test loss, perplexity, and accuracy were $1.77$, $5.86$, and $0.47$, respectively. These results indicate that the proposed derivative-free training procedure can train a nontrivial multi-layer Transformer to nontrivial language-modeling performance without backpropagation, as shown in Table~\ref{tab:generated_sample}.

\begin{table*}[t]
\centering
\caption{Representative character-level text generation result of the MCA-trained Transformer on Tiny Shakespeare.}
\label{tab:generated_sample}
\renewcommand{\arraystretch}{1.15}
\begin{tabular}{p{0.96\textwidth}}
\hline
\textbf{Prompt:}
\\[0.3em]
{\ttfamily\footnotesize
... BIONDELLO:\newline
Forgot you! no, sir: I could not forget you, for I\newline
nev...
}
\\
\hline
\textbf{Generated continuation:}
\\[0.3em]
{\ttfamily\footnotesize
er will made to so hand and speak and it.\newline
\newline
MENENIUS:\newline
What with my reather, reads; I ak to the death many\newline
Clares to with our chame to be my prove\newline
As the we ding from and the hast father of distruck and of my love\newline
Thy shousage of this but their with the sent this and to muchorge\newline
The will for his and breast at you have prouchile\newline
To wing this forth make the arms, and since\newline
my prother sabely, make it.
}
\\
\hline
\end{tabular}
\end{table*}

Although the generated continuation still contains pseudo-words, repeated high-frequency tokens, and local grammatical errors, it reproduces recognizable Shakespeare-like dialogue structure, including speaker labels, line breaks, and dramatic utterance-like segments. Notably, the character-level model continues the truncated prompt across the prompt-generation boundary, extending `nev'' to `never''. This indicates that the MCA-trained Transformer has captured nontrivial character-level regularities and stylistic patterns of the corpus despite being trained without gradient information.

\section{Theoretical Analysis: Comparison between MCA and BP}

BP is a first-order optimization method based on a differentiable loss function and local gradient propagation, whereas MCA is a more general training mechanism based on finite perturbations, comparison of the performance before and after a mutation, and selection rules. In the smooth, small-perturbation limit, the average evolution of MCA can degenerate into local dynamics similar to gradient descent; however, in situations involving non-smooth objectives, discrete weights, hard constraints, local minima, and deep gradient degradation, MCA remains directly applicable. Therefore, MCA should be viewed not as a replacement for BP, but as a complementary approach with greater generality in its optimization mechanism.

\subsection{From Local Gradient to Finite Mutation Selection}

Let the training set be $\mathcal{D} = \{(x_{\mu},y_{\mu})\}_{\mu=1}^{M}$, the neural network be $f_{W}:x\mapsto f_{W}(x)$, and the set of all weights be $W\in\mathbb{R}^{P}$. The empirical risk is written as
\begin{equation}
\mathcal{L}(W) = \frac{1}{M}\sum_{\mu=1}^{M} \ell\left(f_{W}(x_{\mu}),y_{\mu}\right) + \mathcal{R}(W).
\end{equation}
where $\mathcal{R}(W)$ denotes a regularization term (e.g., weight decay). The basic update of BP is
\begin{equation}
W_{t+1} = W_{t} - \eta\nabla_{W}\mathcal{L}(W_{t}),
\end{equation}
where $\eta$ is the learning rate. If parameter constraints exist (e.g., a hard constraint set on the parameters), one usually uses the projected gradient form
\begin{equation}
W_{t+1} = \Pi_{\mathcal{C}}\left(W_{t} - \eta\nabla_{W}\mathcal{L}(W_{t})\right),
\end{equation}
where $\mathcal{C}$ denotes the feasible set for constrained optimization (a specific hard constraint set $\Omega$ on weight magnitudes will be introduced later in this section).

Thus, the core object of BP is the local gradient $\nabla_{W}\mathcal{L}$, which requires that the network mapping, loss function, and the training process as a whole be amenable to chain-rule differentiation. MCA first generates a random mutation
\begin{equation}
W' = W + \Delta W,
\end{equation}
and then decides whether to accept based on the loss change $\Delta\mathcal{L}=\mathcal{L}(W+\Delta W)-\mathcal{L}(W)$. The greedy selection rule can be written as
\begin{equation}
W_{t+1} = \left\{ \begin{array}{ll}
W_{t} + \Delta W, & \Delta\mathcal{L}\leq 0,\\[4pt]
W_{t}, & \Delta\mathcal{L}>0.
\end{array} \right.
\end{equation}

A Metropolis-type acceptance probability can also be used for theoretical analysis of escape behavior:
\begin{equation}
A(W\rightarrow W+\Delta W) = \min\{1,\exp[-\beta\Delta\mathcal{L}]\},
\end{equation}
where $\beta$ denotes the selection intensity. In the experimental part of this paper, we use the greedy rule for simplicity; the Metropolis rule is introduced here solely for theoretical discussion. Therefore, BP relies on local differential information, while MCA relies on finite performance differences. This difference constitutes the fundamental distinction between their capabilities.

\subsection{MCA Contains Gradient Descent in the Local Limit}

Consider the small-perturbation limit for a smooth loss function. Let $g=\nabla_{W}\mathcal{L}(W)$, and assume the perturbation follows a symmetric distribution, e.g., $\Delta W\sim\mathcal{N}(0,\sigma^{2}I)$. For sufficiently small $\sigma$, the loss change satisfies
\begin{equation}
\Delta\mathcal{L}=g\cdot\Delta W + O(\|\Delta W\|^{2}).
\end{equation}

For the Metropolis-type selection rule with symmetric perturbations, the average accepted mutation is proportional to the negative gradient in the small-perturbation limit:
\begin{equation}
\mathbb{E}[\Delta W_{\mathrm{acc}}]\propto -\nabla_{W}\mathcal{L}(W).
\end{equation}
Here the expectation is taken over accepted mutations; the acceptance criterion breaks the symmetry of the proposal distribution, biasing the average toward the negative gradient direction. Thus, the average evolution direction of MCA is consistent with the negative gradient direction. In other words, in the limit of smooth objectives, small perturbations, and strong local selection, MCA can degenerate into dynamics similar to BP or gradient descent. However, the converse does not hold: BP can only use local first-order derivatives, while MCA can also use finite-amplitude mutations and nonlocal selection rules. Hence, BP can be regarded as a special form of mutation--selection dynamics in the locally differentiable limit, whereas MCA retains greater search freedom.

\subsection{Local Minima, Saddle Points, and Advantages of Finite Mutations}

The loss landscape of deep networks typically contains many saddle points, flat regions, and local minima. If a point $W_{*}$ satisfies $\nabla_{W}\mathcal{L}(W_{*})=0$, the first-order update of BP disappears and training tends to stagnate. MCA judges not the gradient but the performance change after a finite perturbation. Expanding to second order around $W_{*}$ gives
\begin{equation}
\Delta\mathcal{L} = \frac{1}{2}\Delta W^{T}H(W_{*})\Delta W + O(\|\Delta W\|^{3}),
\end{equation}
where $H(W_{*})=\nabla_{W}^{2}\mathcal{L}(W_{*})$. If $W_{*}$ is a saddle point, there exists a negative eigenvalue $\lambda_{\min}<0$. Taking $\Delta W = \epsilon v$ along the corresponding eigenvector $v$ yields
\begin{equation}
\Delta\mathcal{L} = \frac{1}{2}\epsilon^{2}\lambda_{\min} < 0.
\end{equation}
This mutation will be accepted by greedy MCA, and will also be accepted with high probability by the Metropolis-type rule. Even near a local minimum, the Metropolis-type acceptance rule (or, more generally, any selection rule with nonzero probability of accepting detrimental mutations) allows the algorithm to accept uphill mutations with nonzero probability, thereby crossing finite energy barriers. This mechanism provides a theoretical ability to escape that differs from BP.

\subsection{Non-smooth Objectives, Discrete Weights, and Hard Constraints}

The applicability of BP relies on the chain rule. When the network contains step functions, sign functions, discrete weights, or non-differentiable structures, the gradient may be zero, non-existent, or fail to reflect true performance changes. BP usually requires surrogate gradients, straight-through estimators, or continuous relaxations. MCA only requires the ability to evaluate $\mathcal{L}(W+\Delta W)-\mathcal{L}(W)$, and does not require $\partial\mathcal{L}/\partial W_{ij}$ to exist. Therefore, it can directly act on discrete or strongly constrained parameter spaces such as
\begin{equation}
W_{ij}\in\{-1,+1\},\quad W_{ij}\in\{-1,0,+1\},\quad |W_{ij}|\leq W_{\max}.
\end{equation}
This means that BP usually optimizes a differentiable relaxation, whereas MCA can search directly in the original design space. This makes MCA particularly relevant for network designs that are low-bit, sparse, discrete, or subject to physical constraints.

In binary synaptic networks, gradient-based backpropagation is not directly applicable; early studies have shown that correlated random walkers with local search strategies can cooperatively find solutions in the discrete weight space, suggesting that mutation--selection dynamics can be effective even when gradient information is absent~\cite{Huang2011}.

\subsection{Weight Constraints and Control of Generalization Ability}

MCA can incorporate structural constraints as part of the search space itself, rather than indirectly penalizing them only through regularization terms. Let
\begin{equation}
\Omega = \{W: |W_{ij}^{(l)}|\leq c_{l},\ \forall i,j,l\}.
\end{equation}
As long as each mutation satisfies $W+\Delta W\in\Omega$, the training process strictly maintains the weight amplitude constraints. There is an analytical connection between this constraint and generalization ability. For a deep network
\begin{equation}
f_{W}(x) = W^{(L)}\phi\left(W^{(L-1)}\cdots\phi(W^{(1)}x)\right),
\end{equation}
if the Lipschitz constant of the activation function is $L_{\phi}$, then
\begin{equation}
\operatorname{Lip}(f_{W}) \leq \prod_{l=1}^{L} L_{\phi} \|W^{(l)}\|_{2}.
\end{equation}
If $|W_{ij}^{(l)}|\leq c_{l}$, then
\begin{equation}
\|W^{(l)}\|_{2} \leq \|W^{(l)}\|_{F} \leq c_{l}\sqrt{n_{l}n_{l-1}},
\end{equation}
so that
\begin{equation}
\operatorname{Lip}(f_{W}) \leq \prod_{l=1}^{L} L_{\phi} c_{l}\sqrt{n_{l}n_{l-1}}.
\end{equation}
Here $n_{l}$ is the number of neurons in layer $l$. This shows that directly constraining the weight amplitude limits the complexity of the network function and provides a basis for controlling generalization. BP usually indirectly controls the parameter scale through soft regularization terms such as $\mathcal{L}_{\text{reg}}(W)=\mathcal{L}(W)+\lambda\|W\|^{2}$, but soft penalties are not strictly equivalent to hard constraints. In contrast, MCA can directly search within a capacity-controlled hypothesis space, and is therefore more suitable for combining with sparsification, discretization, and structural constraint design.

\subsection{Limitations of Gradient Propagation in Deep Networks}

For deep networks, the error signal of BP needs to be propagated backward layer by layer. Let $z^{(l)}=W^{(l)}a^{(l-1)}$, $a^{(l)}=\phi(z^{(l)})$, then the error signal satisfies
\begin{equation}
\delta^{(l)} = (W^{(l+1)})^{T} D^{(l+1)}\delta^{(l+1)},\qquad D^{(l)}=\operatorname{diag}(\phi'(z^{(l)})).
\end{equation}
Thus we have the estimate
\begin{equation}
\|\delta^{(l)}\| \leq \|\delta^{(L)}\| \prod_{r=l+1}^{L} \|W^{(r)}\|_{2} \|D^{(r)}\|_{2}.
\end{equation}
When this product is much less than 1, the gradients in early layers vanish; when it is much greater than 1, the gradients explode. This is a structural difficulty of deep BP training. MCA does not need to backpropagate $\delta^{(l)}$, but instead evaluates the actual output performance change caused by a mutation through forward evaluation. Although the influence of any single weight mutation in a deep network still passes through the forward mapping, MCA uses finite functional changes rather than local differential signals, and thus can theoretically avoid the direct dependence on long chain products of derivatives in BP.

\section{Conclusion and Discussion}

We have demonstrated that the simplest Monte Carlo mutation--optimization selection (MCA) method, implemented on a GPU, can serve as a practical and flexible training paradigm for deep neural networks. Unlike BP, MCA operates without gradient computation or backward error propagation, thereby avoiding gradient degradation issues such as vanishing and exploding gradients. Even with single-parameter random mutations, MCA achieves performance on MNIST comparable to or better than BP for fully connected networks under the full-batch training setting, and exhibits remarkable tolerance to unconventional transfer functions such as the Gaussian function, which are traditionally unfriendly to BP. The flexibility of MCA is further manifested in several aspects that point to its practical value. It enables training through pure pruning, directly removing weights without pre-training a dense network, and reveals substantial redundancy in deep networks. This training-guided sparsification strategy, which contrasts with traditional prune-and-retrain approaches~\cite{Han2015}, offers a new path for model compression and low-power inference, as sparsification can significantly reduce computational load, memory footprint, and energy consumption~\cite{Tmamna2024,Pierro2025,Huang2024}. MCA also naturally supports discrete-weight networks, providing a direct route to hardware-friendly quantization and compatibility with emerging devices such as memristors~\cite{Zhang2025,Aguirre2024}. We have also verified the feasibility of MCA on a simple Transformer architecture for both MNIST image classification and Tiny Shakespeare character-level language modeling, suggesting its potential applicability beyond fully connected networks and beyond classification tasks.

Beyond being a training algorithm, MCA provides a traceable mutation--selection dynamics that offers a new lens for studying the self-organization and learning mechanisms of neural networks---each accepted mutation can be directly linked to performance improvement, sparsification, or representation reorganization. This distinguishes it from the continuous gradient-flow perspective typically associated with BP, and may complement existing theoretical frameworks for understanding generalization, grokking, and the formation of sparse subnetworks~\cite{Frankle2019,Liu2022,Lin2025}.

Regarding training efficiency, MCA exhibits different behaviors across architectures. For fully connected networks, MCA is slower than BP but remains within the same order of magnitude. For single-hidden-layer wide networks, the training time is almost independent of width---a stark contrast to BP, whose per-epoch computational cost scales quadratically with width for a fixed depth. For Transformer architectures, however, MCA is considerably slower than BP. This is the core limitation of MCA. Nevertheless, the current implementation represents only a preliminary optimization of the algorithm, leaving considerable room for further improvement. On the other hand, emerging hardware such as in-memory computing chips provides a naturally friendly platform for MCA-type algorithms. The development of such hardware is expected to substantially accelerate MCA training, making it a more practical approach for training deep networks and even Transformer architectures.

\section*{Acknowledgments}

This work was supported by the National Natural Science Foundation of China (Grant Nos. 11975189 and 12247106). We thank Yuchen Lin, Weichen Fu, and Yong Zhang for useful discussions.

\section*{Conflict of Interest}

The authors declare that they have no conflict of interest.

\bibliographystyle{unsrt}

\end{document}